\documentclass[11pt,a4paper]{article}
\usepackage[hyperref]{naaclhlt2018}
\usepackage{times}
\usepackage{latexsym}
\usepackage{hyperref}
\usepackage{url}
\usepackage{amsmath}
\usepackage{amssymb}
\usepackage{algorithm}
\usepackage{graphicx} 
\usepackage{subfigure}
\graphicspath{ {Figure/} }
\usepackage[noend]{algpseudocode}
\usepackage{multirow}
\usepackage{wrapfig}
\usepackage{lipsum}
\usepackage{float}
\usepackage{url}

\aclfinalcopy 

\title{Embedding Syntax and Semantics of Prepositions via Tensor Decomposition}

\author{Hongyu Gong, Suma Bhat, Pramod Viswanath\\
\{hgong6, spbhat2, pramodv\}@illinois.edu
\\
Department of Electrical and Computer Engineering\\ 
University of Illinois at Urbana-Champaign, USA
}

\date{}

\begin{document}

\maketitle

\begin{abstract}
Prepositions are among the most frequent words in English and play complex roles in the syntax and semantics of sentences. Not surprisingly, they pose well-known difficulties in automatic processing of sentences (prepositional attachment ambiguities and idiosyncratic uses in phrases). 
Existing methods on preposition representation  treat prepositions no different from content words (e.g., word2vec and GloVe). In addition, recent studies aiming at solving prepositional attachment and preposition selection problems
 depend heavily on external linguistic resources and use dataset-specific word representations. 
In this paper we use {\em word-triple} counts (one of the triples being a preposition) to  capture a preposition's interaction with its attachment and complement. We then derive preposition embeddings via tensor decomposition on a large unlabeled corpus.  
We reveal a new geometry involving Hadamard products and empirically demonstrate its utility in paraphrasing  phrasal verbs. Furthermore, our preposition  embeddings are used as simple features in two challenging downstream tasks: preposition selection and prepositional attachment disambiguation. We achieve results comparable to or better than the state-of-the-art on  multiple standardized datasets.  
\end{abstract}

\section{Introduction}
Prepositions are a linguistically closed class comprising some of the most frequent words;  they play an important role in the English language since  they encode rich syntactic and semantic information. Many preposition-related tasks are challenging in computational linguistics because of their polysemous nature and flexible usage patterns. An accurate understanding and representation of prepositions' linguistic role is key to several important NLP tasks such  as grammatical error correction and prepositional phrase attachment. A first-order approach is to represent prepositions as real-valued vectors via word embeddings such as word2vec \cite{mikolov2013efficient} and GloVe \cite{pennington2014glove}. 

Word embeddings  have brought a renaissance in NLP research; they have been very successful in capturing word similarities as well as analogies (both syntactic and semantic) and are now mainstream in nearly all downstream NLP tasks (such as question-answering \cite{DBLP:conf/acl/ChenFWB17}). Despite this success, available literature does not highlight any specific properties of word embeddings of prepositions. Indeed, many of the common prepositions have very similar vector representations as shown in  Table~\ref{tab:prep_sim} for preposition vectors trained using word2vec and GloVe (Tensor embedding is our proposed representation for prepositions). While this suggests that using available representations for prepositions diminishes the distinguishing aspect between prepositions, one could hypothesize that this is primarily because standard word embedding algorithms treat  prepositions no different from other content words such as verbs and nouns, i.e.,  embeddings are created based on  co-occurrences with other words. However, prepositions are very frequent and co-occur with nearly all words, which means that their co-occurrence ought to be treated differently.

\begin{table}[htbp!]
\centering
\label{tab:prep_sim}
\resizebox{0.45\textwidth}{!}{
\begin{tabular}{|c|c|c|c|}
\hline
Preposition pair & Word2vec & GloVe & Tensor \\ \hline
(above, below) & 0.85 & 0.78 & 0.22 \\ \hline
(above, beneath) & 0.40 & 0.45 & 0.15 \\ \hline
(after, before) & 0.83 & 0.70 & 0.44 \\ \hline
(after, during) & 0.56 & 0.42 & 0.16 \\ \hline
(amid, despite) & 0.47 & 0.37 & 0.12 \\ \hline
(amongst, besides) & 0.46 & 0.37 & 0.21 \\ \hline
(beneath, inside) & 0.55 & 0.47 & 0.29 \\ \hline
\end{tabular}}
\caption{Cosine similarity between pairs of centered prepositions using some word embeddings}
\end{table}

Modern descriptive linguistic theory proposes to understand a preposition via  its interactions  with {\em both} the head it attaches to (termed \textit{head}) and its complement \cite{huddleston1984introduction,decarrico2000structure}.
This theory naturally suggests that one should count co-occurrences of a given preposition with {\em pairs} of neighboring words.  One way of achieving this would be by considering a {\em tensor}  of triples ($word_1$, $word_2$, preposition), where  we do not restrict $word_1$ and $word_2$ to be the head and complement words; instead we model a preposition's interaction with {\em all} pairs of  neighboring words via a {\em slice} of a tensor $X$, where the slice is populated by word co-occurrences restricted to a context window of the specific preposition. Thus, the tensor dimension is $N\times N \times K$ where $N$ is the vocabulary size and $K$ is the number of prepositions; since $K\approx 50$, we note  that $N \gg K$. 

Using such a representation, we notice that the resulting tensor is low rank and use it to extract embeddings for both preposition and non-preposition words.  In doing so, we use a combination of standard ideas from word  representations (such as weighted spectral decomposition as in GloVe \cite{pennington2014glove}) and tensor decompositions (alternating least squares (ALS) methods \cite{sharan2017orthogonalized}). We find that the preposition embeddings extracted in this manner are discriminative (see the preposition similarity of the tensor embedding in Table~\ref{tab:prep_sim}). 
Note that the smaller the cosine similarity is, the more distinct the representations are from each other.
We demonstrate that the resulting preposition representation captures the core linguistic properties of prepositions--the attachment and the complement properties. Using both intrinsic evaluations and downstream tasks,  we  show this by providing new state-of-the-art results on well-known NLP tasks involving prepositions. 

\noindent\textbf{Intrinsic evaluations}: We show that the {\em Hadamard} product of the embeddings of a verb and a preposition that together make a phrasal verb, closely approximates the representation of this phrasal verb's paraphrase as a single verb. Example: $v_{\rm made} \odot v_{\rm from} \approx v_{\rm produced} \odot v$,  where $\odot$ represents the Hadamard product (i.e., elementwise multiplication) of two vectors and $v$ is a constant vector (not associated with a specific word and is defined later); this approximation validates that prepositional semantics are appropriately encoded into their trained embeddings. 
We provide a mathematical interpretation for this new geometry while empirically demonstrating the paraphrasing of compositional phrasal verbs.

\noindent\textbf{Extrinsic evaluations}: Our preposition embeddings are used as features for a simple classifier in two well-known  challenging downstream NLP classification tasks. In both tasks, we perform as well as or strictly better than the state-of-the-art on multiple standardized datasets. 

\noindent{\em  Preposition selection}:  While the context in which a preposition occurs governs the choice of the preposition, the specific preposition by itself  significantly influences  the semantics of the context in which it occurs. Furthermore, the choice of the right preposition for a given context can be  very subtle. This idiosyncratic behavior of prepositions  is the reason behind preposition errors being one of the most frequent error types made by second language English speakers \cite{leacock2010automated}).  We demonstrate the utility of the preposition embeddings in the preposition selection task, which is to choose the correct preposition to a given sentence. We show this for a large set of contexts--$7,000$ combined instances from the CoNLL-2013 and the SE datasets \cite{prokofyev2014correct}. Our approach achieves $6\%$ and $2\%$ absolute improvement  over the previous state-of-the-art results on the respective datasets.

\noindent{\em  Prepositional phrase attachment disambiguation}: Prepositional phrase attachment is a common cause of structural ambiguity in natural language. In the sentence ``Pierre Vinken {\em joined the board as a voting member}'', the prepositional phrase ``as a voting member'' can attach to either ``joined'' (the VP) or ``the board'' (the NP); in this case the VP attachment is correct.  
Despite being extensively studied over decades,  prepositional attachment continues to be a major source of syntactic parsing errors \cite{brill1994rule,kummerfeld2012parser, de2016transition}. We use our prepositional representations as simple features to a standard classifier on this task. Our approach tested on a widely studied standard dataset \cite{belinkov2014exploring} achieves 89\% accuracy and compares favorably with the state-of-the art.  
It is noteworthy that while the state-of-the-art results are obtained with  significant  linguistic resources, including syntactic parsers and the WordNet, our approach achieves a comparable performance without relying on such resources. 


We emphasize two aspects of our contributions:\\
(1) Word representations trained via {\em pairwise} word counts are previously shown to capture much of the benefits of the unlabeled sentence-data; 
example: \cite{sharan2017orthogonalized}  reports that their word  representations via word-triple counts  are better than others, but still significantly worse than regular word2vec representations. 
One of our main observations is that considering word-triple counts makes most (linguistic) sense when one of the words is a  preposition. Furthermore, the sparsity of the corresponding tensor is no worse than the sparsity of the regular word co-occurrence matrix (since prepositions are so frequent and co-occur with essentially every word). Taken together, these two points  strongly suggest the benefits of tensor representations in the context for prepositions. \\
(2) The word and preposition representations via tensor decomposition are simple features leading to a standard classifier.  
In particular, we do not use  dependency parsing (which many  prior methods have relied on) or handcrafted features \cite{prokofyev2014correct} or train task-specific representations on the annotated training dataset \cite{belinkov2014exploring}. The simplicity of our approach, combined with the strong empirical results, lends credence to the strength of the prepositional representations found via tensor decompositions.

\section{Method}
\label{sec:method}
We begin with a description of how the tensor with triples (word, word, preposition) is formed and empirically show that its slices are low-rank. Next, we derive low dimensional vector representations for words and prepositions via appropriate tensor decomposition methods.
\begin{figure}[htbp!]
\begin{center}
\includegraphics[width=\linewidth]{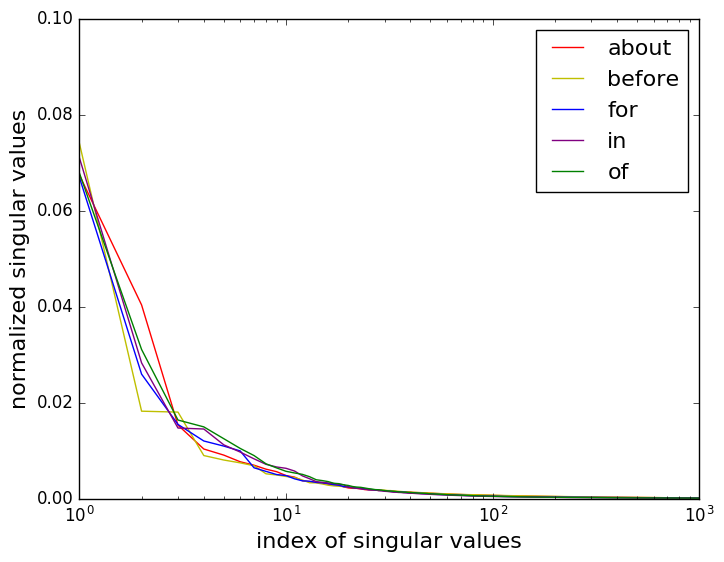}
\caption{Decaying normalized singular values of slices.}
\label{fig:eigen}
\end{center}
\end{figure}

\noindent{\bf Tensor creation}: Suppose that $K$ prepositions are in the preposition set $P=\{p_{1}, \ldots, p_{K}\}$; here $K$ is 49 in our preposition selection task, and 76 in the attachment disambiguation task. We limited the number of prepositions to what was needed in the dataset. 
The vocabulary, the set of all words excluding the prepositions, contains $N$ words, $V=\{w_{1},\ldots,w_{N}\}$, and $N \approx 1M$.   We generate a third order tensor ${\bf X}_{N\times N\times (K+1)}$ from the WikiCorpus \cite{polyglot:2013:ACL-CoNLL} as follows. We say two words co-occur if they appear within a distance $t$ of each other in a sentence. For $k \leq K$, the entry  ${\bf X}_{ijk}$ is the number of occurrences where  word $w_{i}$ co-occurs with preposition $p_k$, and $w_{j}$ also co-occurs with preposition $p_k$ in the same sentence, and this is counted  across all sentences in the WikiCorpus. 
For $0 \leq k \leq K$, ${\bf X}[:,:,k]$ is a matrix of the count of the word pairs that co-occur with the preposition $k$, and we call such a matrix a slice. 

Here we use a window of size $t=3$. While prepositions co-occur with many words, 
there are also a number of other words which do not occur in the context of any preposition. In order to 
make the maximal use  of the data, we add an extra slice ${\bf X}[:,:,K+1]$, where the entry ${\bf X}_{ij(K+1)}$ is the number of occurrences where $w_i$ co-occurs with $w_j$ (within distance $2t = 6$) but at least one of them is not within a distance of $t$ of any  preposition. Note that the preposition window of $3$ is smaller than the word window of $6$, since it is known that the interaction between prepositions and neighboring words usually weakens more sharply with distance when compared to that of content words \cite{hassani2017disambiguating}.

\noindent{\bf Empirical properties of} ${\bf X}$: We find that the tensor $X$ is very sparse -- only $1\%$ of the tensor elements are non-zero. Furthermore,  $\log(1+{\bf X}[:,:,k])$ is low-rank (here the logarithm is applied component-wise to every entry of the tensor slice). Towards seeing this, we choose slices  corresponding to the prepositions ``about'', ``before'',``for'', ``in'' and ``of'', and plot their normalized singular values in Figure~\ref{fig:eigen}. We see that the singular values decay dramatically, suggesting the low-rank structure in each slice.


\noindent{\bf Tensor decomposition}: We combine standard ideas from word embedding algorithms and tensor decomposition algorithms to arrive at the low-rank approximation to the tensor $\log(1+{\bf X})$. In particular, we consider two separate methods: 

1. {\em Alternating Least Squares (ALS)}. A generic method to decompose a tensor into its modes is via the  CANDECOMP/PARAFAC (CP) decomposition \cite{kolda2009tensor}.  The tensor $\log(1+{\bf X})$ is decomposed into three modes: ${\bf U}_{d\times N}$, ${\bf W}_{d\times N}$ and ${\bf Q}_{d\times (K+1)}$, based on  the solutions to the  optimization problem (\ref{eq:als}).  Here   ${\bf u}_{i}$, ${\bf w}_{i}$ and ${\bf q}_{i}$ are the $i$-th column of $U$, $W$ and $Q$, respectively.
\begin{align}
\label{eq:als}
\nonumber
L &= \min\limits_{{\bf U},{\bf W},{\bf Q}}
\sum\limits_{i=1}^{N}\sum\limits_{j=1}^{N}\sum\limits_{k=1}^{K+1}\left(\right.\langle {\bf u}_{i}, {\bf w}_{j}, {\bf q}_{k} \rangle \\
&- \log (1+ {\bf X}_{ijk}) \left)\right.^{2},
\end{align}
where $\langle {\bf a}, {\bf b}, {\bf c} \rangle= {\bf 1}^t ({\bf a} \odot {\bf b} \odot {\bf c})$ is the inner product of three vectors ${\bf a}, {\bf b}$ and ${\bf c}$. Here ${\bf 1}$ is the column vector of all ones and $\odot$ refers to the Hadamard product. 
We can interpret the columns of $U$ as the word representations and the columns of $Q$ as the preposition representations, each of dimension $d$ (equal to 200 in this paper). There are several algorithmic solutions to this optimization problem in the literature, most of which are based on alternating least squares methods \cite{kolda2009tensor,comon2009tensor,anandkumar2014guaranteed} and we employ a recent one named Orth-ALS \cite{sharan2017orthogonalized} in this paper. Orth-ALS  periodically orthogonalizes the decomposed components while fixing two modes and updating the remaining one. 
It is supported by theoretical guarantees and empirically outperforms standard ALS methods in different applications.

2. {\em Weighted Decomposition (WD)}: Based on ideas from the literature on word embedding algorithms, we also consider  weighting different elements of the tensors differently in order to reduce the effect of the large dynamic range of the tensor values. Specifically, we employ the  GloVe objective function to our tensor model and  minimize the  objective function (\ref{eq:wd}):
\begin{align}
\label{eq:wd}
\nonumber
&L_{\text{weighted}} = \min\limits_{{\bf U},{\bf W},{\bf Q}}\sum\limits_{i=1}^{N}\sum\limits_{j=1}^{N}\sum\limits_{k=1}^{K+1}\omega_{ijk}\left(\right.\langle {\bf u}_{i}, {\bf w}_{j}, {\bf q}_{k} \rangle \\
&+ b_{Ui} + b_{Wj} + b_{Qk} - \log({\bf X}_{ijk}+1) \left)\right.^{2},
\end{align}
where $b_{Ui}$ is the scalar bias for  the word $i$ in the matrix $U$. Similarly, $b_{Wj}$ is the bias for the word $j$ in the matrix $W$, and $b_{Qk}$ for preposition $k$ in the matrix $Q$. Bias terms are learned in such a way as to minimize the loss function. Here $\omega_{ijk}$ is the weight assigned to each tensor element $X_{ijk}$, and we use the weighting proposed by GloVe: $$\omega_{ijk}=\min\left(\left(\frac{{\bf X}_{ijk}}{x_{\text{max}}}\right)^{\alpha}, 1\right).$$ 
We set the hyperparameters  to be $x_{\text{max}}=10$, and $\alpha=0.75$ in this work. 
We solve this optimization problem via standard gradient descent, arriving at word representations ${\bf U}$ and tensor representations ${\bf Q}$.

\begin{table*}[htbp!]
\centering
\label{tab:para}
\begin{tabular}{|c|c|c|c|c|c|c|}
\hline 
{\bf Phrase} & replied to & blocked off & put in & pray for & dreamed of & sparked off\\ \hline
{\bf Paraphrase} & answered & intercepted & place & hope & wanted  & prompted  \\ \hline
{\bf Phrase} & stuck with & derived from & switched over & asked for & passed down & blend in \\ \hline
{\bf Paraphrase} & stalled & generated & transferred & requested & delivered  & mix \\ \hline
\end{tabular}
\caption{Paraphrasing of prepositional phrases.}
\end{table*}

\section{Geometry of Phrasal Verbs}
\label{sec:geometry}
{\bf  Representation Interpretation}
Suppose that we have a  phrase $(h,p_{i},c)$ where $h$, $p_{i}$ and $c$ are the head word, the preposition $i (i\le K)$ and the complement respectively. The inner product of the word vectors of $h,p_{i}$ and $c$ reflects how frequently $h$ and $c$ co-occur in the context of $p$. It also reflects how  cohesive  the triple is. 

Recall that there is an extra $(K+1)-$th slice that describes the word co-occurrences outside the preposition window, which considers cases such as the verb phrase $(v, c)$ where $v$ and $c$ are the verb and its complement without a preposition in their shared context.   
Now consider a phrasal verb \textit{sparked off} and a verb phrase with head  \textit{prompted}.
For any complement word $c$ that fits these two phrases--the phrasal verb having $h$ as its  head verb and $p_i$ as its preposition, and the other, the verb phrase with $v$ as  its head--
we can expect that
\begin{align*}
\langle {\bf u}_{h}, {\bf q}_{i}, {\bf w}_{c} \rangle \approx \langle {\bf u}_{v}, {\bf q}_{K+1}, {\bf w}_{c} \rangle.
\end{align*}
In other words 
${\bf u}_{h}\odot {\bf q}_{i} \approx {\bf u}_{v}\odot {\bf q}_{K+1}
$,  where ${\bf a} \odot {\bf b}$ denotes the pointwise multiplication (Hadamard product) of vectors ${\bf a}$ and ${\bf b}$. 
This suggests that: (1) The vector  ${\bf q}_{K+1}$ is a constant vector for all $(v, c)$ pairs, and that (2) we could paraphrase the verb phrase $(h,p_{i})$ by finding a verb $v$ such that ${\bf u}_{v}\odot {\bf q}_{K+1}$ is closest to ${\bf u}_{h}\odot {\bf q}_{i}$. 
\begin{align}
\label{eq:paraphrase}
{\rm paraphrase} = \arg\min\limits_{v} \lVert {\bf u}_{v}\odot {\bf q}_{K+1}- {\bf u}_{h}\odot {\bf q}_{i}\rVert.
\end{align} 
This shows that well-trained embeddings are able to capture the relation between phrasal verbs  and their equivalent single verb forms. 

In Table~\ref{tab:para}, we list paraphrases of some verb phrases, which are generated from the weighted tensor decomposition. As can be seen, the tensor embedding gives reasonable paraphrasing, which validates that the trained embedding is interpretable in terms of lexical semantics.




In the next two sections, we evaluate the proposed tensor-based preposition embeddings in the context of two important NLP downstream tasks: preposition selection and preposition attachment disambiguation.
In this work, we use the English WikiCorpus (around 9 GB) as the training corpus for different sets of embeddings. We train tensor embeddings with both Orth-ALS and weighted decomposition.  The implementation of Orth-ALS is built upon the SPLATT toolkit \cite{splattsoftware}. 
We perform orthogonalization in the first $5$ iterations in  Orth-ALS decomposition, and the training is completed when its performance stabilizes. As for the weighted decomposition, we train for $20$ iterations, and its hyperparameters are set as $x_{\text{max}}=10$, and $\alpha=0.75$.

We also include two baselines for comparison--word2vec's CBOW model and GloVe. We set $20$ training iterations for both the models. The hyperparameters in word2vec are set as: window size=6, negative sampling=25 and down-sampling=1e-4.
The hyperparameters in GloVe are set as: window size=6, $x_{max}$=10, $\alpha$=0.75 and minimum word count=5. We note that all the representations in this study--word2vec, GloVe and our tensor embedding--are of dimension 200.

\section{Downstream Application: Preposition Selection}
\label{sec:exp}

Grammatical error detection and correction constitute important tasks in NLP. Among grammatical errors, prepositional errors constitute about $13\%$ of all errors, ranking second among the most common error types \cite{leacock2010automated}. This is due to the fact that prepositions are highly polysemous and have idiosyncratic usage. Selecting a preposition depends on how well we can capture the interaction between a preposition and its context. Hence we choose this task to  evaluate how well the lexical interactions are captured by different methods. 

\begin{table}[htbp!]
\centering
\resizebox{0.35\textwidth}{!}{
\begin{tabular}{|c|c|c|}
\hline
\multirow{3}{*}{FCE} & \# of sent & 27119 \\ \cline{2-3} 
 & \# of prep & 60279 \\ \cline{2-3} 
 & Error ratio & 4.8 \\ \hline
\multirow{3}{*}{CoNLL} & \# of sent & 1375 \\ \cline{2-3} 
 & \# of prep & 3241 \\ \cline{2-3} 
 & Error ratio & 4.7 \\ \hline
\multirow{3}{*}{SE} & \# of sent & 5917 \\ \cline{2-3} 
 & \# of prep & 15814 \\ \cline{2-3} 
 & Error ratio & 38.2 \\ \hline
\end{tabular}
}
\caption{Dataset statistics.}
\label{tab:PrepSelectionData}
\end{table}

\textbf{Task}. Given a sentence in English containing a preposition, we either replace the preposition with the correct one or retain it. For example, in the sentence ``It can save the effort to carrying a lot of cards,''  ``to'' should be corrected as ``of.''  Formally, there is a closed set of preposition candidates $P=\{p_{1}, \ldots, p_{m}\}$. A preposition $p$ is used in a sentence $s$ consisting of words $s=\{\ldots, w_{-2}, w_{-1}, p, w_{1}, w_{2},\ldots\}$. If used incorrectly, we need to replace $p$ by another preposition $\hat{p}\in P$ based on the context.

\begin{table*}[htbp!]
\centering
\resizebox{0.65\textwidth}{!}{
\begin{tabular}{|c|c|c|c|c|}
\hline
Dataset & Method  & Precision & Recall & F1 score \\ \hline
\multirow{5}{*}{CoNLL} & State-of-the-art & 0.2592 & 0.3611 & 0.3017 \\ \cline{2-5}
 & Word2vec & 0.1558 & 0.1579 & 0.1569 \\ \cline{2-5}
 & GloVe & 0.1538 & 0.1578 & 0.1558 \\ \cline{2-5}
 & Our method (ALS) & 0.3355 & 0.3355 & 0.3355 \\ \cline{2-5}
 & Our method  (WD) & 0.3590 & 0.3684 & {\bf 0.3636} \\ \hline
\multirow{5}{*}{SE} & State-of-the-art & 0.2704 & 0.2961 & 0.2824 \\ \cline{2-5}
 & Word2vec & 0.2450 & 0.2585 & 0.2516 \\ \cline{2-5}
 & GloVe & 0.2454 & 0.2589 & 0.2520 \\ \cline{2-5}
 & Our method (ALS) & 0.2958 & 0.3146 & {\bf 0.3049} \\ \cline{2-5}
 & Our method (WD) & 0.2899 & 0.3055 & 0.2975 \\ \hline
\end{tabular}}
\caption{Performance on preposition selection.}
\label{tab:prepSelectRes}
\end{table*}


\textbf{Dataset}. For training, we use the data from the Cambridge First Certificate in English (FCE) exam, just as used by the state-of-the-art on preposition error correction \cite{prokofyev2014correct}. As for test data, we use two the CoNLL-2013 and the Stack Exchange (SE) datasets. The CoNLL dataset on preposition error correction was published by the CoNLL 2013 shared task \cite{ng2014conll}, collected from 50 essays written by 25 non-native English learners at a university. The SE dataset consists of texts generated by
non-native speakers on the Stack Exchange website. Detailed  statistics are shown in Table~\ref{tab:PrepSelectionData}.
We focus on the most frequent $49$ prepositions listed in Appendix~\ref{app:roster}.

\textbf{Evaluation metric}. Three metrics--precision, recall and F1 score--are used to evaluate the preposition selection performance. 

\textbf{Our algorithm}. We first preprocess the dataset by removing articles, determiners and pronouns, and take a context window of $3$. We divide the task into two steps: error detection and error correction. Firstly, we decide whether a preposition is used correctly in the context. If not, we suggest another preposition as replacement in the second step. The detection step uses only three features: the cosine similarity between the the current preposition embedding and the average context embedding, the rank of the preposition in terms of this  cosine similarity, and the probability that this preposition is not changed in the training corpus. We build a decision tree classifier with these three features and find that we can identify errors with $98\%$ F1 score in the CoNLL dataset  and $96\%$ in the SE dataset.

For the error correction part, we only focus on the errors detected in the first stage. Suppose that the original preposition is $q$, and the candidate preposition is $p$ with the embedding $v_{p}$. The word vectors in the left context window are averaged as the left context embedding $v_{\ell}$, and the right vectors are averaged to give the right context embedding $v_{r}$. We have the following features:
\begin{enumerate}
\item  Embedding features: ${\bf v}_{\ell}, {\bf v}_{p}$ and ${\bf v}_{r}$;
\item Pair similarity between the preposition and the context:  maximum of the similarity of the preposition between the left and the right context, i.e.,
$\text{pair~sim}=\max\left(\frac{{\bf v}_{\ell}^{T}{\bf v}_{p}}{\lVert {\bf v}_{\ell}\rVert_{2}\cdot\lVert {\bf v}_{p}\rVert_{2}}, \frac{{\bf v}_{r}^{T}{\bf v}_{p}}{\lVert {\bf v}_{r}\rVert_{2}\cdot\lVert {\bf v}_{p}\rVert}_{2}\right)$; 
\item $\text{Triple~similarity} =\frac{\langle {\bf v}_{\ell},{\bf v}_{p},{\bf v}_{r}\rangle}{\lVert {\bf v}_{\ell}\rVert_{3}\cdot \lVert {\bf v}_{p}\rVert_{3}\cdot\lVert {\bf v}_{r}\rVert_{3}}$;
\item Confusion probability: the probability that $q$ is replaced by $p$ in the training data.
\end{enumerate}

A two-layer feed-forward neural network (FNN) with hidden layer sizes of 500 and 10 is trained with these features to score prepositions in each sentence. The preposition with the highest score is the suggested edit.

\textbf{Baseline}. The state-of-the-art on preposition selection uses n-gram statistics from a  large corpus \cite{prokofyev2014correct}. Features such as point-wise mutual information (PMI) and part-of-speech tags are fed into a supervised scoring system. Given a sentence with a preposition to either replace or retain, the preposition with the highest score is chosen. 

The performance of the baseline is affected by both the system architecture and the features. To evaluate the benefits brought about by our tensor embedding-based features, we also consider other baselines which have the same two-step architecture whereas the features are generated from word2vec and GloVe embeddings. These baselines allow us to compare the representation power independent  of the classifier. 

\begin{table*}[htbp!]
\centering 
\resizebox{\textwidth}{!}{
\begin{tabular}{|c|c|c|c|c|c|c|c|}
\hline
\multicolumn{2}{|c|}{\begin{tabular}[c]{@{}c@{}}Removed\\ feature\end{tabular}} & \begin{tabular}[c]{@{}c@{}}Left context \\ embedding\end{tabular} & \begin{tabular}[c]{@{}c@{}}Prep\\ embedding\end{tabular} & \begin{tabular}[c]{@{}c@{}}Right context\\ embedding\end{tabular} & \begin{tabular}[c]{@{}c@{}}Pair\\ similarity\end{tabular} & \begin{tabular}[c]{@{}c@{}}Triple\\ similarity\end{tabular} & \begin{tabular}[c]{@{}c@{}}Confusion \\ score\end{tabular} \\ \hline
\multirow{3}{*}{CoNLL} & Precision & 0.1558 & 0.2662 & 0.3117 & 0.3247 & 0.3247 & 0.3506 \\ \cline{2-8} 
 & Recall & 0.1579 & 0.2697 & 0.3158 & 0.3289 & 0.3289 & 0.3553 \\ \cline{2-8} 
 & F1 score & 0.1569 & 0.2680 & 0.3137 & 0.3268 & 0.3268 & 0.3529 \\ \hline
\multirow{3}{*}{SE} & Precision & 0.2587 & 0.2796 & 0.2649 & 0.2658 & 0.2647 & 0.1993 \\ \cline{2-8} 
 & Recall & 0.2743 & 0.2964 & 0.2801 & 0.2818 & 0.2807 & 0.2114 \\ \cline{2-8} 
 & F1 score & 0.2663 & 0.2877 & 0.2726 & 0.2735 & 0.2725 & 0.2052 \\ \hline
\end{tabular}}
\caption{Ablation analysis in preposition selection.}
\label{tab:ablationPrepSelect}
\end{table*}

\textbf{Result}. We compare our proposed embedding-based method against baselines mentioned in Table~\ref{tab:prepSelectRes}. We note that the proposed tensor embeddings achieve the best performance among all approaches. In particular, the tensor with weighted decomposition has the highest F1 score on the CoNLL dataset--a $6\%$ improvement over the state-of-the-art. However, the tensor with ALS decomposition performs the best on the SE dataset, achieving a $2\%$ improvement over the state-of-the art. 
We also note that with the same architecture, tensor embeddings perform much better than word2vec and GloVe embeddings on both the datasets. This validates the representation power of tensor embeddings of prepositions.

To get a deeper insight into the importance of the features in the preposition selection task, we also performed an ablation analysis of the tensor method with weighted decomposition as shown in Table~\ref{tab:ablationPrepSelect}. 
 We find that the left context is the most important feature in for the CoNLL dataset, whereas the confusion score is the most important for the SE dataset. Pair similarity and triple similarity are less important when compared with the other features. This is because the neural network was able to learn the lexical similarity from the embedding features, thus reducing the importance of the similarity features.

\textbf{Discussion}. Now we analyze different cases where our approach selects the wrong preposition.
(1) \emph{Limited context window}.
We focus on the local context within  a preposition's window. In some cases, we find that head words might be out of the context window. 
An instance of this is found in the sentence ``prevent more of this kind of tragedy \emph{to} happening''  \emph{to} should be corrected as \emph{from}. Given the context window of $3$, we cannot get the lexical clues provided by \emph{prevent}, which leads to the selection error. (2) \emph{Preposition selection requires more context}.
Even when the context window contains all the words on which the preposition depends, it still may not be sufficient to select the right one. For example, in the sentence ``it is controlled by some men \emph{in} a bad purpose'' where our approach replaces the preposition \emph{in} with the preposition \emph{on} given the high frequency of the phrase ``on purpose''. The correct preposition should be \emph{for} based on the whole sentence. 

\section{Downstream Application: Prepositional Attachment}
In this section, we discuss the task of prepositional phrase (PP) attachment disambiguation, a well-studied, but hard task in syntactic parsing. 
The PP attachment disambiguation inherently requires an accurate description of the interactions among the head, the preposition and the complement, which becomes an ideal task to evaluate our tensor-based embeddings.

\textbf{Task}. The English dataset used in this work is collected from a linguistic treebank by \cite{belinkov2014exploring}. It provides $35,359$ training and $1,951$ test instances. 
Each instance consists of several head candidates, a preposition and a complement word. The task is to pick the head to which the preposition attaches. In the example ``he saw an elephant with long tusks'', the words ``saw'' and ``elephant'' are the  candidate head words.


\textbf{Our algorithm}. Let ${\bf v}_{h}, {\bf v}_{p}$ and ${\bf v}_{c}$ be embeddings for the head candidate $h$, preposition $p$ and child $c$ respectively. We then use the following features:
\begin{enumerate}
\item Embedding feature: candidate head, preposition and complement embedding;
\item Triple similarity: $\frac{\langle {\bf v}_{h}, {\bf v}_{p}, {\bf v}_{c}\rangle}{\lVert {\bf v}_{h}\rVert_{3}\cdot\lVert {\bf v}_{p}\rVert_{3}\cdot\lVert {\bf v}_c\rVert_{3}}$;
\item Head-preposition similarity: $\frac{{\bf v}_h^{T}{\bf v}_{p}}{\lVert {\bf v}_h\rVert_{2}\cdot\lVert {\bf v}_p\rVert_{2}\cdot}$;
\item Head-child similarity: $\frac{{\bf v}_h^{T}{\bf v}_c}{\lVert {\bf v}_h\rVert_{2}\cdot\lVert {\bf v}_c\rVert_{2}\cdot}$;
\item Part-of-speech (pos) tag of candidates and next words;
\item Distance between $h$ and $p$.
\end{enumerate}

We use a basic neural network, a two-layer feed-forward network (FNN) with hidden-layers of size $1000$ and $20$, to take the input features and predict the probability that a candidate is the head. The candidate with the highest likelihood is chosen as the head.

\begin{table*}[htbp!]
\centering
\resizebox{\textwidth}{!}{
\begin{tabular}{|c|c|c|c|c|c|c|c|}
\hline
Classifier & \begin{tabular}[c]{@{}c@{}}HPCD \\ (enriching)\end{tabular} & LRFR & OntoLSTM & FNN & FNN & FNN & FNN \\ \hline
\begin{tabular}[c]{@{}c@{}}Embedding\\ method\end{tabular} & GloVe & Word2vec & \begin{tabular}[c]{@{}c@{}}Glove-\\ extended\end{tabular} & Word2vec & GloVe & \begin{tabular}[c]{@{}c@{}}Our method \\ (ALS)\end{tabular} & \begin{tabular}[c]{@{}c@{}}Our method \\ (WD)\end{tabular} \\ \hline
Resources & \begin{tabular}[c]{@{}c@{}}POS tag, \\ WordNet, \\ VerbNet \end{tabular} & \begin{tabular}[c]{@{}c@{}}POS tag, \\ WordNet, \\ VerbNet\end{tabular} & \begin{tabular}[c]{@{}c@{}}POS tag, \\ WordNet\end{tabular} & POS tag & POS tag & POS tag & POS tag \\ \hline
Accuracy & 0.887 & {\bf 0.903} & 0.897 & 0.866 & 0.858 & 0.883 & 0.892 \\ \hline
\end{tabular}}
\caption{Accuracy in prepositional attachment disambiguation.}
\label{tab:prepAttachRes}
\end{table*}

\textbf{Baselines}. For comparison, we include the following state-of-the-art approaches in preposition attachment disambiguation. The linguistic resources they used to enrich their features are listed in Table~\ref{tab:prepAttachRes}. \\
(1) Head-Prep-Child-Dist (HPCD) Model \cite{belinkov2014exploring}: this compositional neural network is used to train task-specific representations of prepositions. \\
(2) Low-Rank Feature Representation (LRFR) \cite{yu2016embedding}: this method incorporates word parts, contexts and labels into a tensor, and uses decomposed vectors as features for disambiguation.\\
(3) Ontology LSTM (OntoLSTM) \cite{dasigi2017ontology}:  the vectors are initialized with GloVe, extended by AutoExtend \cite{rothe2015autoextend}, and trained via LSTMs for head selection.

Similar to the experiments in the preposition selection task (see Section 4), we also include baselines which have the same feed-forward network architecture but generate features with vectors trained by word2vec and GloVe. 
They are denoted as FNN with different initializations in Table~\ref{tab:prepAttachRes}. 
Since the attachment disambiguation is a selection task,  accuracy is a natural evaluation metric. 
\begin{table*}[htbp!]
\centering
\resizebox{\textwidth}{!}{
\begin{tabular}{|c|c|c|c|c|c|c|c|c|}
\hline
\begin{tabular}[c]{@{}c@{}}Removed \\ feature\end{tabular} & \begin{tabular}[c]{@{}c@{}}Head \\ vector\end{tabular} & \begin{tabular}[c]{@{}c@{}}Prep\\ vector\end{tabular} & \begin{tabular}[c]{@{}c@{}}Child \\ vector\end{tabular} & \begin{tabular}[c]{@{}c@{}}Head-prep\\ similarity\end{tabular} & \begin{tabular}[c]{@{}c@{}}Head-child\\ similarity\end{tabular} & \begin{tabular}[c]{@{}c@{}}Triple\\ similarity\end{tabular} & POS & Distance \\ \hline
Accuracy & 0.843 & 0.871 & 0.880 & 0.877 & 0.885 & 0.873 & 0.850 & 0.872 \\ \hline
\end{tabular}}
\caption{Ablation analysis in preposition attachment disambiguation.}
\label{tab:ablationPrepAttach}
\end{table*}

\textbf{Result}. We compare the results of the different approaches and the linguistic resources used in Table~\ref{tab:prepAttachRes}, where we see that our simple classifier built on the tensor representation is comparable in performance to the state-of-the-art (within 1\% of the result). This result is notable considering that prior competitive approaches have used significant linguistic resources such as VerbNet and WordNet, whereas we use none. With the same feed-forward neural network as the classifier, our tensor-based approaches (both ALS and WD) achieve better performance than word2vec and GloVe.

An ablation analysis that is provided in Table~\ref{tab:ablationPrepAttach} shows that the head vector feature affects the performance the most (indicating that  heads interact more closely with prepositions), and the POS tag feature comes second. 
 The similarity features appear less important since the classifier has access to the lexical relatedness via the embedding features. 
 Prior works have reported the importance of  the distance feature  since $81.7\%$ sentences take the word closest to the preposition as the head. In our experiments, the distance feature was found to be less important compared to the embedding features.

\noindent {\bf Discussion}.  We found that one source of attachment disambiguation error is the lack of a broader context in our features. 
A broader context is critical in  examples such as ``worked'' and ``system,'' which are head candidates of ``for trades'' in the sentence ``worked on a system for trading''. 
They are reasonable heads in the expressions ``worked for trades'' and ``system for trades'' and further disambiguation requires a context larger than what we considered.


\section{Related Work}
\textbf{Word representation}. Word embeddings have been successfully used in many NLP applications. Word2vec \cite{mikolov2013efficient} and GloVe \cite{pennington2014glove} show that embeddings can capture lexical semantics very well. \citet{zhang2014word} studied embeddings which can generalize different similarity perspectives when combined with corresponding linear transformations. Unlike other words, the crucial syntactic roles of prepositions in addition to their rich semantic meanings have been highlighted in prior works \cite{hovy2010,schneider2015hierarchy}. Nevertheless, word  representations specifically focused on prepositions are not available and to the best of our knowledge, ours is the first work exploring this intriguing direction. 

\textbf{Tensor Decomposition}. Tensors embed higher order interaction among different modes, and the tensor decomposition  captures this  interaction via lower dimensional representations.  There are several decomposition methods such as
Alternating Least Square (ALS) \cite{kolda2009tensor}, Simultaneous Diagonalization (SD) \cite{kuleshov2015tensor} and optimization-based methods \cite{liu1989limited,more1978levenberg}. Orthogonalized Alternating Least Square (Orth-ALS) adds the step of component orthogonalization to each update step in the ALS method \cite{sharan2017orthogonalized}. Owing to its theoretical  guarantees and, more relevantly due to its good empirical performance, Orth-ALS  is the algorithm of choice in this paper.

\textbf{Preposition Selection}. Preposition selection, an important area of study in  computational linguistics, is also a very practical topic in the context of grammar correction and second language learning. Prior works have used hand-crafted heuristic rules \cite{xiang2013hybrid}, n-gram features  \cite{prokofyev2014correct,rozovskaya2013university}, and by the use of POS tags and dependency relations to enrich other features  \cite{kao2013conll}--all toward addressing preposition error correction.

\textbf{Prepositional Attachment Disambiguation}. There is a storied literature on prepositional attachment disambiguation, long recognized as  an important part of syntactic parsing \cite{kiperwasser2016simple}. Recent works, based on word embeddings have pushed the boundary of state of the art empirical results. A seminal work in this direction is the Head-Prep-Child-Dist Model, which trained embeddings in a compositional network to maximize the accuracy of head prediction \cite{belinkov2014exploring}. The performance has been further improved in conjunction with semantic and syntactic features. 
A recent work has  proposed an initialization with semantics-enriched GloVe embeddings, and retrained representations with LSTM-RNNs \cite{dasigi2017ontology}. Another recent work has used tensor decompositions to capture the relation between word representations and their labels \cite{yu2016embedding}.

\section*{Conclusion}
\label{sec:conclusion}
Co-occurrence counts of word pairs in sentences and the resulting word vector representations (embeddings) have revolutionalized NLP research. A natural generalization is to consider co-occurrence counts of word triples, resulting in a third order tensor. Partly due to the size of the tensor (a vocabulary of 1M, leads to a tensor with $10^{18}$ entries!) and partly due to the extreme dynamic range of entries (including sparsity), word vector representations via tensor decompositions have largely been inferior to their lower order cousins (i.e., regular word embeddings). 

In this work, we trek this well-trodden but arduous terrain by restricting word triples to the scenario when one of the words is a preposition. This is  linguistically justified, since prepositions are understood to model interactions between pairs of words. Numerically, this is also very well justified since the sparsity and dynamic range of the resulting tensor is no worse than the original matrix of pairwise co-occurrence counts; this is because prepositions are very frequent and co-occur with essentially every word in the vocabulary. 

Our intrinsic evaluations and new state-of-the-art results in downstream evaluations   lend strong credence to the tensor-based  approach to prepositional representation. We expect our vector representations of prepositions to be widely used in more complicated downstream NLP tasks where prepositional role is crucial, including ``text to programs'' \cite{guu2017language}.


\bibliography{naaclhlt2018}

\begin{thebibliography}{}
\expandafter\ifx\csname natexlab\endcsname\relax\def\natexlab#1{#1}\fi

\bibitem[{Al-Rfou et~al.(2013)Al-Rfou, Perozzi, and
  Skiena}]{polyglot:2013:ACL-CoNLL}
Rami Al-Rfou, Bryan Perozzi, and Steven Skiena. 2013.
\newblock \href{http://www.aclweb.org/anthology/W13-3520}{Polyglot: Distributed
  word representations for multilingual nlp}.
\newblock In {\em Proceedings of the Seventeenth Conference on Computational
  Natural Language Learning\/}. Association for Computational Linguistics,
  Sofia, Bulgaria, pages 183--192.
\newblock \url{http://www.aclweb.org/anthology/W13-3520}.

\bibitem[{Anandkumar et~al.(2014)Anandkumar, Ge, and
  Janzamin}]{anandkumar2014guaranteed}
Animashree Anandkumar, Rong Ge, and Majid Janzamin. 2014.
\newblock Guaranteed non-orthogonal tensor decomposition via alternating
  rank-$1 $ updates.
\newblock {\em arXiv preprint arXiv:1402.5180\/} .

\bibitem[{Belinkov et~al.(2015)Belinkov, Lei, Barzilay, and
  Globerson}]{belinkov2014exploring}
Yonatan Belinkov, Tao Lei, Regina Barzilay, and Amir Globerson. 2015.
\newblock Erratum: "exploring compositional architectures and word vector
  representations for prepositional phrase attachment".
\newblock {\em {TACL}\/} 3:101.

\bibitem[{Brill and Resnik(1994)}]{brill1994rule}
Eric Brill and Philip Resnik. 1994.
\newblock A rule-based approach to prepositional phrase attachment
  disambiguation.
\newblock In {\em Proceedings of the 15th conference on Computational
  linguistics-Volume 2\/}. Association for Computational Linguistics, pages
  1198--1204.

\bibitem[{Chen et~al.(2017)Chen, Fisch, Weston, and
  Bordes}]{DBLP:conf/acl/ChenFWB17}
Danqi Chen, Adam Fisch, Jason Weston, and Antoine Bordes. 2017.
\newblock \href{https://doi.org/10.18653/v1/P17-1171}{Reading wikipedia to
  answer open-domain questions}.
\newblock In {\em Proceedings of the 55th Annual Meeting of the Association for
  Computational Linguistics, {ACL} 2017, Vancouver, Canada, July 30 - August 4,
  Volume 1: Long Papers\/}. pages 1870--1879.
\newblock \url{https://doi.org/10.18653/v1/P17-1171}.

\bibitem[{Comon et~al.(2009)Comon, Luciani, and De~Almeida}]{comon2009tensor}
Pierre Comon, Xavier Luciani, and Andr{\'e}~LF De~Almeida. 2009.
\newblock Tensor decompositions, alternating least squares and other tales.
\newblock {\em Journal of chemometrics\/} 23(7-8):393--405.

\bibitem[{Dasigi et~al.(2017)Dasigi, Ammar, Dyer, and
  Hovy}]{dasigi2017ontology}
Pradeep Dasigi, Waleed Ammar, Chris Dyer, and Eduard Hovy. 2017.
\newblock Ontology-aware token embeddings for prepositional phrase attachment.
\newblock {\em arXiv preprint arXiv:1705.02925\/} .

\bibitem[{de~Kok and Hinrichs(2016)}]{de2016transition}
Dani{\"e}l de~Kok and Erhard Hinrichs. 2016.
\newblock Transition-based dependency parsing with topological fields.
\newblock In {\em Proceedings of the 54th Annual Meeting of the Association for
  Computational Linguistics\/}. volume~2, pages 1--7.

\bibitem[{DeCarrico(2000)}]{decarrico2000structure}
Jeanette~S DeCarrico. 2000.
\newblock {\em The structure of English: Studies in form and function for
  language teaching\/}, volume~1.
\newblock University of Michigan Press/ESL.

\bibitem[{Guu et~al.(2017)Guu, Pasupat, Liu, and Liang}]{guu2017language}
Kelvin Guu, Panupong Pasupat, Evan~Zheran Liu, and Percy Liang. 2017.
\newblock From language to programs: Bridging reinforcement learning and
  maximum marginal likelihood.
\newblock {\em arXiv preprint arXiv:1704.07926\/} .

\bibitem[{Hassani and Lee(2017)}]{hassani2017disambiguating}
Kaveh Hassani and Won-Sook Lee. 2017.
\newblock Disambiguating spatial prepositions using deep convolutional
  networks.
\newblock In {\em AAAI\/}. pages 3209--3215.

\bibitem[{Hovy et~al.(2010)Hovy, Tratz, and Hovy}]{hovy2010}
Dirk Hovy, Stephen Tratz, and Eduard Hovy. 2010.
\newblock What's in a preposition?: dimensions of sense disambiguation for an
  interesting word class.
\newblock In {\em Proceedings of the 23rd International Conference on
  Computational Linguistics: Posters\/}. Association for Computational
  Linguistics, pages 454--462.

\bibitem[{Huddleston(1984)}]{huddleston1984introduction}
Rodney Huddleston. 1984.
\newblock {\em Introduction to the Grammar of English\/}.
\newblock Cambridge University Press.

\bibitem[{Kao et~al.(2013)Kao, Chang, Chiu, Yen, Boisson, Wu, and
  Chang}]{kao2013conll}
Ting-Hui Kao, Yu-Wei Chang, Hsun-Wen Chiu, Tzu-Hsi Yen, Joanne Boisson,
  Jian-Cheng Wu, and Jason~S Chang. 2013.
\newblock Conll-2013 shared task: Grammatical error correction nthu system
  description.
\newblock In {\em CoNLL Shared Task\/}. pages 20--25.

\bibitem[{Kiperwasser and Goldberg(2016)}]{kiperwasser2016simple}
Eliyahu Kiperwasser and Yoav Goldberg. 2016.
\newblock Simple and accurate dependency parsing using bidirectional lstm
  feature representations.
\newblock {\em arXiv preprint arXiv:1603.04351\/} .

\bibitem[{Kolda and Bader(2009)}]{kolda2009tensor}
Tamara~G Kolda and Brett~W Bader. 2009.
\newblock Tensor decompositions and applications.
\newblock {\em SIAM review\/} 51(3):455--500.

\bibitem[{Kuleshov et~al.(2015)Kuleshov, Chaganty, and
  Liang}]{kuleshov2015tensor}
Volodymyr Kuleshov, Arun Chaganty, and Percy Liang. 2015.
\newblock Tensor factorization via matrix factorization.
\newblock In {\em Artificial Intelligence and Statistics\/}. pages 507--516.

\bibitem[{Kummerfeld et~al.(2012)Kummerfeld, Hall, Curran, and
  Klein}]{kummerfeld2012parser}
Jonathan~K Kummerfeld, David Hall, James~R Curran, and Dan Klein. 2012.
\newblock Parser showdown at the wall street corral: An empirical investigation
  of error types in parser output.
\newblock In {\em Proceedings of the 2012 Joint Conference on Empirical Methods
  in Natural Language Processing and Computational Natural Language
  Learning\/}. Association for Computational Linguistics, pages 1048--1059.

\bibitem[{Leacock et~al.(2010)Leacock, Chodorow, Gamon, and
  Tetreault}]{leacock2010automated}
Claudia Leacock, Martin Chodorow, Michael Gamon, and Joel Tetreault. 2010.
\newblock Automated grammatical error detection for language learners.
\newblock {\em Synthesis lectures on human language technologies\/}
  3(1):1--134.

\bibitem[{Liu and Nocedal(1989)}]{liu1989limited}
Dong~C Liu and Jorge Nocedal. 1989.
\newblock On the limited memory bfgs method for large scale optimization.
\newblock {\em Mathematical programming\/} 45(1):503--528.

\bibitem[{Mikolov et~al.(2013)Mikolov, Chen, Corrado, and
  Dean}]{mikolov2013efficient}
Tomas Mikolov, Kai Chen, Greg Corrado, and Jeffrey Dean. 2013.
\newblock Efficient estimation of word representations in vector space.
\newblock {\em arXiv preprint arXiv:1301.3781\/} .

\bibitem[{Mor{\'e}(1978)}]{more1978levenberg}
Jorge~J Mor{\'e}. 1978.
\newblock The levenberg-marquardt algorithm: implementation and theory.
\newblock In {\em Numerical analysis\/}, Springer, pages 105--116.

\bibitem[{Ng et~al.(2014)Ng, Wu, Briscoe, Hadiwinoto, Susanto, and
  Bryant}]{ng2014conll}
Hwee~Tou Ng, Siew~Mei Wu, Ted Briscoe, Christian Hadiwinoto, Raymond~Hendy
  Susanto, and Christopher Bryant. 2014.
\newblock The conll-2014 shared task on grammatical error correction.
\newblock In {\em CoNLL Shared Task\/}. pages 1--14.

\bibitem[{Pennington et~al.(2014)Pennington, Socher, and
  Manning}]{pennington2014glove}
Jeffrey Pennington, Richard Socher, and Christopher Manning. 2014.
\newblock Glove: Global vectors for word representation.
\newblock In {\em Proceedings of the 2014 conference on empirical methods in
  natural language processing (EMNLP)\/}. pages 1532--1543.

\bibitem[{Prokofyev et~al.(2014)Prokofyev, Mavlyutov, Grund, Demartini, and
  Cudr{\'e}-Mauroux}]{prokofyev2014correct}
Roman Prokofyev, Ruslan Mavlyutov, Martin Grund, Gianluca Demartini, and
  Philippe Cudr{\'e}-Mauroux. 2014.
\newblock Correct me if i'm wrong: Fixing grammatical errors by preposition
  ranking.
\newblock In {\em Proceedings of the 23rd ACM International Conference on
  Conference on Information and Knowledge Management\/}. ACM, pages 331--340.

\bibitem[{Rothe and Sch{\"u}tze(2015)}]{rothe2015autoextend}
Sascha Rothe and Hinrich Sch{\"u}tze. 2015.
\newblock Autoextend: Extending word embeddings to embeddings for synsets and
  lexemes.
\newblock {\em arXiv preprint arXiv:1507.01127\/} .

\bibitem[{Rozovskaya et~al.(2013)Rozovskaya, Chang, Sammons, and
  Roth}]{rozovskaya2013university}
Alla Rozovskaya, Kai-Wei Chang, Mark Sammons, and Dan Roth. 2013.
\newblock The university of illinois system in the conll-2013 shared task.
\newblock In {\em Proceedings of the Seventeenth Conference on Computational
  Natural Language Learning: Shared Task\/}. pages 13--19.

\bibitem[{Schneider et~al.(2015)Schneider, Srikumar, Hwang, and
  Palmer}]{schneider2015hierarchy}
Nathan Schneider, Vivek Srikumar, Jena~D Hwang, and Martha Palmer. 2015.
\newblock A hierarchy with, of, and for preposition supersenses.
\newblock In {\em Proceedings of The 9th Linguistic Annotation Workshop\/}.
  pages 112--123.

\bibitem[{Sharan and Valiant(2017)}]{sharan2017orthogonalized}
Vatsal Sharan and Gregory Valiant. 2017.
\newblock Orthogonalized als: A theoretically principled tensor decomposition
  algorithm for practical use.
\newblock {\em arXiv preprint arXiv:1703.01804\/} .

\bibitem[{Smith and Karypis(2016)}]{splattsoftware}
Shaden Smith and George Karypis. 2016.
\newblock {SPLATT: The Surprisingly ParalleL spArse Tensor Toolkit}.
\newblock \url{http://cs.umn.edu/~splatt/}.

\bibitem[{Xiang et~al.(2013)Xiang, Yuan, Zhang, Wang, Zheng, and
  Wei}]{xiang2013hybrid}
Yang Xiang, Bo~Yuan, Yaoyun Zhang, Xiaolong Wang, Wen Zheng, and Chongqiang
  Wei. 2013.
\newblock A hybrid model for grammatical error correction.
\newblock In {\em CoNLL Shared Task\/}. pages 115--122.

\bibitem[{Yu et~al.(2016)Yu, Dredze, Arora, and Gormley}]{yu2016embedding}
Mo~Yu, Mark Dredze, Raman Arora, and Matthew Gormley. 2016.
\newblock Embedding lexical features via low-rank tensors.
\newblock {\em arXiv preprint arXiv:1604.00461\/} .

\bibitem[{Zhang et~al.(2014)Zhang, Salwen, Glass, and Gliozzo}]{zhang2014word}
Jingwei Zhang, Jeremy Salwen, Michael Glass, and Alfio Gliozzo. 2014.
\newblock Word semantic representations using bayesian probabilistic tensor
  factorization.
\newblock In {\em Proceedings of the 2014 Conference on Empirical Methods in
  Natural Language Processing (EMNLP)\/}. pages 1522--1531.

\end{thebibliography}
\bibliographystyle{acl_natbib}

\appendix
\section{Roster of Prepositions}
\label{app:roster}
The list of most frequent 49 Prepositions in the task of preposition selection is shown below:

about, above, absent, across, after, against, along, alongside,
amid, among, amongst, around, at, before, behind,
below, beneath, beside, besides, between, beyond, but, by,
despite, during, except, for, from, in, inside, into, of, off,
on, onto, opposite, outside, over, since, than, through, to,
toward, towards, under, underneath, until, upon, with.

\end{document}